# Training Tactile Sensors to Learn Force Sensing from Each Other

## (First Online at March 2, 2025)

**Authors:**

Zhuo Chen[1*], Ni Ou[1], Xuyang Zhang[1], Zhiyuan Wu[1], Yongqiang Zhao[1], Yupeng Wang[1], Emmanouil Spyrakos Papastavridis[1], Nathan Lepora[2], Lorenzo Jamone[3], Jiankang Deng[4*], Shan Luo[1*]

**Affiliations:**

[1]King's College London, London, United Kingdom.
[2]University of Bristol, Bristol, United Kingdom.
[3]University College London, London, United Kingdom.
[4]Imperial College London, London, United Kingdom.

*Corresponding authors. Email: shan.luo@kcl.ac.uk, zhuo.7.chen@kcl.ac.uk, j.deng16@imperial.ac.uk

**Abstract:** Humans achieve stable and dexterous object manipulation by coordinating grasp forces across multiple fingers and palms, facilitated by a unified tactile memory system in the somatosensory cortex. This system encodes and stores tactile experiences across skin regions, enabling the flexible reuse and transfer of touch information. Inspired by this biological capability, we present GenForce, the first framework that enables transferable force sensing across tactile sensors in robotic hands. GenForce unifies tactile signals into shared marker representations, analogous to cortical sensory encoding, allowing force prediction models trained on one sensor to be transferred to others without the need for exhaustive force data collection. We demonstrate that GenForce generalizes across both homogeneous sensors with varying configurations and heterogeneous sensors with distinct sensing modalities and material properties. This transferable force sensing is also demonstrated with high performance in robot force control including daily object grasping, slip detection and avoidance. Our results highlight a scalable paradigm for cross-sensor robotic tactile learning, offering new pathways toward adaptable and tactile memory-driven manipulation in unstructured environments.

## Main Text:

Robots capable of perceiving and interacting with the environment are revolutionizing healthcare, service, and manufacturing industry[1,2]. While these robots rely on various sensing modalities, tactile sensing is indispensable for achieving dexterous manipulation and ensuring safe human-robot interaction[3,4]. However, replicating the human sense of touch in robotic systems remains a fundamental challenge, particularly for tasks requiring precise in-hand force feedback.

Human skin is equipped with diverse sensory receptors that detect mechanical stimuli and provide rich contact information[5]. Inspired by this, tactile robotics aims to mirror the function of human mechanoreceptors to enhance dexterity and intelligence[6,7] by developing various bio-inspired tactile sensing systems (**Fig. 1**A). These tactile sensors are based on various sensing principles, such as piezoresistive[8], capacitive[9], magnetic[10,11], and optical[12,13] transductions, offering unique advantages in detecting static or dynamic contact information. For example (**Fig. 1**A), GelSight[13] provides high-resolution tactile images by using a camera to capture detailed textures and geometries, making it ideal for tasks requiring fine-grained tactile feedback, like object recognition or 3D reconstruction[14]. uSkin[15] uses Hall-effect sensors to measure displacements of embedded magnets, offering high sensitivity to multi-axis forces and compactness for precise manipulation tasks. TacTip[16,12] is featured by tracking biomimetic pin movement with a camera, offering robust force, shape, and slip detection abilities in dynamic environments.

However, diversity in sensing principles, structural designs, and material properties creates significant domain gaps among tactile sensors, hindering the transfer of learned tactile experience and necessitating repetitive data collection. For example, training force prediction models[17] for robotic hands with multiple tactile sensors (**Fig. 1**B) requires extensive paired tactile signal-force label datasets for each sensor, introducing practical barriers through costly force/torque sensors (such as ATI Nano17) and time-consuming data collection. This challenge is exacerbated by the inevitable degradation of soft elastomers from wear and tear, requiring frequent replacement and recalibration. Thus, current practice demands repeated collection of paired tactile signal-force labels and retraining of force prediction models for each new sensor deployment.

The ability to unify and transfer sensory information across skin regions is a critical aspect of human control over grasp forces, ensuring both stability and dexterity during object manipulation[18,19]. This capability is equally essential for robots equipped with tactile sensors to perform in-hand object manipulation tasks in unstructured environments[20]. In humans, the tactile memory system (**Fig. 1**C) enables the storage and retrieval of experienced tactile information, such as haptic stimuli, across skin regions on hands[21,22]. Mechanoreceptors in the skin detect deformation, which is translated into a unified sensory encoding, and transmitted to the somatosensory cortex via peripheral nerves for storage and processing[5]. This human ability to adapt, unify, and transfer tactile sensation offers valuable inspiration for developing transferable tactile sensing in robots. Mimicking this unified representation and transferable tactile sensing could enable tactile sensors to learn from each other, reuse collected tactile experience and transfer tactile sensing across robot hands, enhancing their dexterity and adaptability.

Recent advances in transfer learning[23] and representation learning[24–28] have made strides toward unifying latent tactile representations and enabling the transfer of tactile experience. However, existing approaches face significant limitations compared to the human capability of tactile memory. For example, traditional transfer learning techniques still require the collection of labeled data from new sensors for finetuning[29]. Recent reported representation learning methods[24–28] focus only on vision-based tactile sensors and align sensor representations in feature space, necessitating large-scale datasets from different sensors and often depend on task-specific decoders trained with labeled data to enhance accuracy. Some efforts to directly transform tactile signals across sensors[30–33] similarly fall short on one-to-one translation, as they overlook the importance of a unified tactile representation among sensors and lack of generalizability to diverse tactile sensors (see

Supplementary Table 1). Critically, the above approaches fail to account for the fundamental material differences among sensors, which are essential for tasks demanding high accuracy such as force prediction.

We present GenForce (**Fig. 1**D), the first general framework to enable transferable force sensing across diverse tactile sensors. GenForce unifies tactile signals from skin deformation into marker representations (**Fig. 1**D), analogous to sensory encoding, and enables deformation transfer across sensors regardless of sensing principles and physical configurations through marker-to-marker translation (**Fig. 2**C). Force prediction capability (**Fig. 2**D) can therefore be learned with generated images and force labels from other sensors. GenForce further improves force accuracy by compensating for material differences. Extensively validated for generalizability, accuracy, and robustness, GenForce works across flat-surface vision-based tactile sensor GelSight [13], flat-surface magnetic sensor arrays uSkin [15] with three-axis sensing or z-axis-only sensing, and curved-surface vision-based tactile sensor TacTip [16], all with varying configurations (**Fig. 2**A-B). This approach significantly redefines the paradigm of training force prediction models: instead of repetitive and time-consuming data collection for each sensor, models can be trained by learning across sensors, greatly facilitating large-scale tactile sensing deployment. Beyond force sensing, GenForce is successfully deployed on real robot tasks with force control, including daily objects grasping, slip detection and avoidance (see Supplementary Video 4-7). Our study lays a foundation for transferable sensing in tactile robotics[34] and paves the way for tactile robots capable of transferring multimodal perception in whole body toward human-like embodied AI in the future.

# RESULTS

## Arbitrary marker-to-marker translation

To test the generalizability of the marker-to-marker translation (M2M) for diverse tactile sensors, we first propose a simple simulation pipeline to acquire extensive deformed marker images (**Fig. 3**A). We design 12 distinct reference marker patterns (**Fig. 3**A), i.e. *Array (A)*, *Circle (C)* and *Diamond (D)* shapes with varying sizes and densities referring to GelSight[13], uSkin[15], TacPalm[16] and GelTip[35] sensors. Eighteen 3D-printed indenters with diverse geometrical properties (vertices, edges, and curvatures) are used for indentation[33]. Each marker pattern can serve as both source sensor and target sensor, resulting in a total of 12×11=132 sensor combinations. We employ two quantitative metrics[36], including Fréchet Inception Distance (FID) and Kernel Inception Distance (KID), to assess image similarity. Lower values indicate greater visual similarity. The visualization of feature spaces uses t-distributed stochastic neighbor embedding (t-SNE)[37]. Note that, we abbreviate each *source-to-target* combination as *source_target*. For example, we write sensor combination *D2-to-C1* as *D2_C1* for simplification.

As demonstrated in **Fig. 3**B-C and Supplementary Video 1, approximately 100-fold decreases are found in FID and KID across all combinations after using M2M, indicating the generated images successfully represent the image styles of the targe images while aligning the deformations with the source images. Specifically, before using M2M (**Fig. 3**B-i and **Fig. 3**C-i), as each type of marker pattern is visually distinctive (**Fig. 3**A), the difference is prominent with an average FID larger than 400 and an average KID larger than 0.75. The source images and target images are separated distinctly in the feature space (**Fig. 3**D-i). After using M2M, the average FID (**Fig. 3**B-ii) drops to 4 and the average KID (see **Fig. 3**C-ii) drops to 0.01. The generated images and target images are aligned closely in feature space (**Fig. 3**D-ii) and visually indistinguishable (Supplementary Figure 1D). This suggests the effectiveness of the marker-to-marker translation across various marker patterns, providing a pretrained model for the M2M translation in real sensors.

## Learning across homogeneous tactile sensors

Large-scale deployment of homogeneous tactile sensors on robotic hands has garnered significant attention[38,39]. A common approach is to directly apply trained force prediction models to other sensors[39]. However, sensor variations often exist due to inconsistencies in fabrication and will lead to large force errors by using such source-only method. We validate our GenForce model across homogeneous sensors, particularly for GelSight sensors due to widespread use in robotic hands[14,38,39]. We detail the study of material effects in the next section and focus here on the variations in illuminations and marker patterns.

We build the data collection setup (**Fig. 4**A) and fabricate five GelSight sensors (**Fig. 4**), i.e. *Array-I (A-I)*, *Array-II (A-II)*, *Circle-I (C-I), Circle-II (C-II)*, and *Diamond-I (D-I)*. We divide the indenters into seen group with 12 indenters and unseen group with 6 indenters (Supplementary Figure 1A). Approximately 180,000 force-image pairs are collected per sensor. For M2M model training, we only use the last four images during the movement to a target position as location-paired images (**Fig. 4**B), resulting in a total of 17,280 images per sensor. The distribution of the data across different force ranges is shown in Supplementary Figure 6A. It covers normal forces from -16 N to 0 N and shear forces from -6 N to 6 N. Collected tactile images are showcased in Supplementary Figure 3. We utilize marker segmentation methods (see Supplementary Figure 4B and Methods) to obtain marker-based tactile images. The pretrained M2M model is then fine-tuned with location-paired marker images contacting with indenters from seen group.

Similar to the simulation results, image similarity (**Fig. 4**C-D) improve significantly and the generated images and the target images across all combinations are well-aligned (Supplementary Figure 5B). The FID and KID values both drop by more than 98%. The results tested only on indenters from unseen group are presented in Supplementary Figure 5C-E, demonstrating comparable performances to the seen group. The generated images are showcased in Supplementary Figure 5Aand Supplementary Video 1, proving the generalizability of the M2M model to real-world homogeneous tactile sensors.

Regarding force prediction performance, the source-only method exhibits large errors (**Fig. 4** and Supplementary Video 2). The maximum error in normal force exceeds 4.8 N. Shear force errors are average above 0.28 N. In addition, most sensor combinations have negative $R^2$ values and high variance (**Fig. 4**) and demonstrate poor performance in real-time force prediction (**Fig. 4**). After using the GenForce model, all force errors are significantly reduced (**Fig. 4**), and $R^2$ values (**Fig. 4**) improve across all combinations. For normal force, the maximum error is below 1 N, while the minimum error decreases to less than 0.7 N. Notably, the maximum force error in *C-II_D-I* improves from 4.8 N to 0.96 N. For shear forces, the minimum errors decrease to less than 0.06 N. The $R^2$ values (**Fig. 4**) show consistent improvement in both normal and shear direction, averaging above 0.8. Real-time force prediction in **Fig. 4** and Supplementary Video 2 demonstrate well-aligned predicted forces and ground truth. The half-violin plots (Supplementary Figure 6B) demonstrate the high performance for normal forces (-8 N to 0 N) and shear forces ( –3 N to 3 N). Additional results tested on unseen group (Supplementary Figure 6C-D) further demonstrate our model's effectiveness and generalizability.

## Skin hardness effect and compensation

The deformable skin of tactile sensors is akin to human skin, where its hardness property is often customized for specific tasks, and it also changes with long-term use and replacement due to wear and tear. The difference in material hardness introduces additional errors when transferring the force labels (**Fig. 5**A). We investigate the skin hardness effect and compensation method in this section. Seven silicone elastomers with varied hardness are fabricated by controlling the base to activator (B:A) ratio (**Fig. 5**) from ratio-6 (*r6*) to ratio-18 (*r18*) with a step of 2. The higher the ratio, the softer the skin becomes. We measure the force-depth curve (**Fig. 5**) for each skin during loading and unloading phases. Due to the hysteresis property, the contact force in the loading phase is found greater than that in the unloading phase at the same depth. This effect reveals the inherent errors of traditional methods to predict forces using single images[17]. The measured relationship in **Fig. 5**D between the shear-to-normal force $F_S / F_N$ and shear displacement reveals distinct friction of coefficients across seven elastomers (see Supplementary Text 9 for material compensation). The Examples of images selected from seven sensors with varying hardness are showcased in Supplementary Figure 7. The distributions of collected points in different force ranges are shown in Supplementary Figure 9A.

We divide the transfer direction into two groups: (1) *hard-to-soft* and (2) *soft-to-hard*. In **Fig.5**E-i, combinations like *6_8, 6_10*, and *6_12* indicate progressively transfer to harder elastomer (slightly harder, harder, much harder), whereas *8_6*, *10_6*, and *12_6* indicate the inverse, i.e., softer transfer. These combinations reflect the long-term aging behavior of silicone elastomers. Before material compensation, normal-force errors generally grow with increasing hardness gap (**Fig. 5**E). On average, the normal-force error is 1.41 N for *hard-to-soft* and 1.03 N for *soft-to-hard*. For shear forces, average errors are 0.18 N (x) and 0.20 N (y) in *hard-to-soft*, and 0.16 N (x) and 0.18 N (y) in *soft-to-hard*. Shear errors are more stable across both groups, because applied shear forces are smaller than normal forces and less sensitive to hardness gaps (see Supplementary Text 9A). Additional results for 15 other translation combinations are provided in Supplementary Fig. 8D–E.

To mitigate the increased force errors due to material hardness gaps, we propose a material compensation method (**Fig. 5**B) that uses material priors to correct force labels before training. The priors are the fitted elastic modulus profiles, i.e force-normalized depth profiles shown in Supplementary Figure 8C. Intuitively, this process increases the amount of force when transferring from soft to hard skin and decreases it in the opposite direction. As shown in **Fig. 5**E-F, this improves accuracy across both normal and shear forces. After compensation, average normal-force error drops to 0.99 N in the *hard-to-soft* group (30% reduction) and 0.87 N in the *soft-to-hard* group (16% reduction). For shear forces, average errors in the *hard-to-soft* group reduce to 0.10 N on the x-axis (44% reduction) and 0.14 N on the y-axis (30% reduction); in the *soft-to-hard* group, errors reduce to 0.13 N ($F_x$, 19%) and 0.15 N ($F_y$, 19%). The method yields lower errors in 95% of combinations within the 21 *hard-to-soft* combinations (including those in Supplementary Fig. 8D) and in 57% of the 21 *soft-to-hard* combinations. A representative case, transferring from r18 (soft) to r6 (much harder) skins in **Fig. 5**G, shows improvements in $R^2$ values from 0.73 ($F_x$), 0.79($F_y$), and 0.78 ( $F_z$ ) to 0.87, 0.86, and 0.84 across the three axes. Real-time visualizations in Supplementary Fig. 9B and Supplementary Video 2 further illustrate the fidelity of the material compensation, which is ignored by traditional studies while common in real-world applications.

## Learning across heterogeneous tactile sensors

Heterogeneous tactile sensors are designed with varying sensing principles, structural designs, and material properties to mimic human mechanoreceptors, offering additional challenges to learn from each other. For example, TacTip[16] with 127 markers with a curvature of 41.5 mm and a diameter of 40 mm, and allows for a maximum indentation of 5 mm. By contrast, GelSight[13] is equipped with 56 markers with a size of 25×25×4 mm and the maximum indentation is 1.5 mm. uSkin[15] has a dimension of 24.6×22.6×5 mm and outputs multichannel signals via 4×4 taxels (16 markers), where each taxel can afford a maximum force of 20N.

We unify three distinct tactile signals (Supplementary Figure 11A) into marker representation. For electronic sensor arrays, we develop a signal-to-marker pipeline (Fig. **6** A and Supplementary Figure 10A-B) that converts multichannel raw signals into marker displacement and diameter change (**Fig. 6**A). An example conversion from full uSkin (3-axis) is provided in Supplementary Fig. 12B. Although some tactile arrays (e.g., capacitive or resistive) can only measure pressure in each taxel compared to magnetic sensor with 3-axis measurement, our model still transfers from vision-based tactile sensors to these arrays. We verify this by using only the z component of uSkin, referred to as "uSkin (z-axis)" in Supplementary Text 10 and Supplementary Video 3. For TacTip, we extract markers using segmentation methods (Supplementary Figure 10C). The measured force-normalized depth relationship for three sensors with varying indentation depth: 1 mm (uSkin), 1 mm (GelSight), and 4.5 mm (TacTip) are shown in Supplementary Figure 13A. Despite TacTip's deeper contact depth, its force at maximum normalized depth is still four times smaller than uSkin and two times smaller than GelSight due to the extremely soft gel used as the soft skin[16]. The surface frictions are measured shown in Supplementary Figure 13B and 14A. Examples of tactile images for three heterogeneous tactile sensors are shown in Supplementary Figure 11B-D. The distribution of data points collected across different force ranges is shown in Supplementary Figure 16B..

The improved FID (**Fig. 6**) and KID (**Fig. 6**) and the aligned feature space in tSNE (Supplementary Figure 12C) suggests the M2M model's generalization to heterogeneous sensors. We showcase generated images and source images in **Fig. 6** and the Supplementary Video 1. For force prediction, we observe that (**Fig. 6**), the $R^2$ values are negative across all three axes when transferring from lower marker density to higher density using source-only method, i.e. *uSkin_TacTip*, *uSkin_GelSight*, and *GelSight_TacTip*. The maximum MAE reaches 7.76 N for $F_z$, 0.59 N for $F_y$, and 0.37 N for $F_x$ . By contrast, from high density to lower density, the remaining three combinations exhibit better performance but still have high errors. These large force errors are also coupled with the hardness effect, which is prominent when transferring from uSkin sensor, with the hardest skin, to the other two sensors with softer skin. After using GenForce but without compensation, force prediction errors (Supplementary Figure 15A-ii) improve substantially. However, due to the remaining gaps in hardness, *GelSight_TacTip* and *uSkin_TacTip* still exhibit negative $R^2$ values (Supplementary Figure 15A-i). Visualization in Supplementary Figure 15Balso demonstrates the gap between predicted force and ground truth due to the hardness gaps.

Upon applying material compensation, all $R^2$ values (**Fig. 6**i) improve to positive values, following an ascending performance order: *GelSight_uSkin*, *uSkin_GelSight*, *TacTip_GelSight*, *GelSight_TacTip*, *TacTip_uSkin*, and *uSkin_TacTip*. This order correlates with material hardness gaps and marker density. GelSight and uSkin, having the smallest hardness and marker density gaps, show the lowest errors compared to combinations involving TacTip. The average of MAE (**Fig. 6**) over all six combinations decreases below 0.92 N for normal force, while $F_x$ and $F_y$ reduce below 0.22 N and 0.3 N, respectively. Notably, the *uSkin_TacTip* group shows a 93% improvement of $F_z$ in MAE (from 7.76 N to 0.52 N) and a 66% improvement of $F_y$ (from 0.59 N to 0.2 N). The continuous force predictions in **Fig. 6** also demonstrate the mitigation of hardness gap. In Supplementary Figure 16C, the force errors for all combinations are centered around zero within ranges of -4N to 0N in normal direction and -3N to 3N in shear direction, demonstrating both the accuracy and reliability of our model in the most challenging task of heterogeneous translation.

We evaluated our model in real-time across six transfer groups under more dynamic conditions (see Supplementary Video 3). Tactile sensors were mounted on an ATI Nano17 F/T sensor, and we applied forces using four daily objects with different shapes and materials, including screwdriver, glue stick, plastic pizza, and a LEGO block (Supplementary Fig. 17). A human operator performed five common dynamic contact events, including press, rub, roll, push, and pull, as well as continuous combinations of these on the sensor surface (Fig. 6G). The transfer groups covered both homogeneous and heterogeneous translations described earlier and also included translation from a vision-based tactile sensor (GelSight, D–I) to a sensor array with z-axis only sensing capability (uSkin, z-axis). As shown in Supplementary Video 3 and Supplementary Figure 18-22, all test groups exhibited fast, accurate responses comparable to a commercial F/T sensor.

## Transferable force sensing and control in robot manipulation

To demonstrate the applicability of our model in real world, we install heterogeneous tactile sensors on a robot arm to show the transferable force sensing and control in daily-object robot grasping, slip detection and avoidance (See Supplementary Figure 25A). The robot setup can be seen in **Fig. 7A-i.** In the robot grasping task, we mount a flat-surface vision-based tactile sensor (GelSight, marker pattern A-II) on the left finger of robot and a flat-surface magnetic tactile sensor (uSkin with three-axis sensing capability) on the right finger. We transfer the force prediction model to above two sensors by using a third flat-surface vision-based tactile sensor (GelSight, marker pattern A-II) tactile sensor with our GenForce model. The task requires robot to grasp nine daily objects (See Supplementary Figure 25B) with different sizes, shapes and material without damage them. Those objects include potato chip, grape, strawberry, orange, plum, wood block, glue stick, meat box and tea box, which are unseen in the training dataset. During the grasping, the arm is controlled with a proportional controller to grasp those objects with fixed normal forces ranging from 0.6N to 1.2N. Both sensors are shared with the same force controller. As shown in **Fig. 7C,** Supplementary Video 4 and Supplementary Video 5, both sensors can equip the robot arm an accurate force sensing so that the robot arm can successfully grasp all objects with target forces without damage. Even for challenging objects such as chips and fresh fruits, the robot can achieve delicate grasping by using the transferred force prediction model combining with force control.

For the second task, we extend the grasping experiment (see **Fig. 7**A-ii; Supplementary Videos 6 and 7) to include slip detection and avoidance. A curved-surface vision-based TacTip sensor (palm shape) is mounted on the left finger, and a three-axis magnetic uSkin sensor is mounted on the right.

The TacTip's force prediction model is transferred from GelSight (D-I), and the uSkin model is subsequently transferred from the TacTip. The task proceeds through several stage: moving down, proportional-control grasping, lifting, slip detection and avoidance at the top position, release, and return to home with the force controller active only during the grasp and slip-detection phases. We evaluate four objects: banana, plum, meat box, and glue stick. Beyond completing the grasp, external forces are applied by a human at the top position to induce slip. The robot detects slip via changes in shear force and responds by narrowing the gripper width. As shown in **Fig. 7B** and Supplementary Videos 6–7, the system completes all stages successfully, demonstrating the practical applications of our model in real robotic tasks.

# DISCUSSION

In summary, this study tackles the fundamental challenge of transferring force sensing capabilities across tactile sensors. While advances inspired by human mechanoreceptors have led to diverse bioinspired tactile sensors, few studies have focused on bridging those sensors or transferring tactile experience between them. This need becomes critical in large-scale tactile sensing systems, where collecting labeled data for each sensor is impractical and costly.

Traditional paradigm for endowing force sensing capability for tactile sensors present significant limitations. For example, Finite Element Method (FEM) approaches[40] rely on sensor-specific stiffness matrices and assume linear deformation of soft elastomers[41]. While learning-based methods[42] better handle nonlinear responses, they require costly force/torque sensors (e.g., Nano17) and extensive data collection for every new sensor. GenForce overcomes these hurdles by mimicking the human tactile memory system to unify, adapt, and transfer tactile experiences, allowing sensors to acquire force prediction capabilities by directly learning from one another.

Unlike recent methods[24–28] that focus solely on vision-based tactile sensors with learned latent representations requiring extensive sensor data and limited to homogeneous sensors, our GenForce model operates directly at the tactile-signal level and is generalizable to a wide variety of tactile sensors. By employing a unified marker representation, analogous to human sensory encoding, our method explicitly transfers skin deformation between sensors through marker-to-marker translation and validates the transferable force sensing in robot manipulation tasks. This provides a transformative paradigm for transferable tactile sensing across tactile sensors.

The key contributions of our study can be summarized as:

1. Cost-efficient and user-friendly: Our approach alleviates the need for expensive force/torque sensors when training force prediction model for new sensors. It only requires few location-paired data, less than 10% of the force-paired images used in our experiments, substantially reducing time and cost.
2. Foundational and Scalable: The marker-to-marker translation, based on an image-conditioned diffusion model, enables transfer of marker representations across a wide variety of sensors using a single model. Once translation is established between sensors, no additional data is needed, making the model inherently scalable as sensor networks grow.
3. Generalizability: Our model generalizes diverse tactile sensors regardless of sensing principles, structural designs, or material properties. This is validated on simulated data with 132 sensor combinations and real-world data with 74 combinations, across both homogeneous and heterogeneous translations.

4. Accuracy and Reliability: Despite utilizing unsupervised learning, the force prediction errors are tightly centered around zero within -4 N to 0 N in normal direction and -3 N to 3 N in shear direction across all transfer combinations. The model is also validated in real-time compared with ATI nano 17 F/T sensor under dynamic contact events commonly in manipulation tasks.
5. Applicability: We successfully demonstrate the model on robot daily-objects grasping, slip detection and avoidance with force control. The trained model is robust to environmental noise arising from sensor fabrication, data collection, marker segmentation, and translation processes.

However, we still find our model shows few failure cases:

1. We observed a flicker effect when transferring to GelSight (A-I) (see Supplementary Fig. 41 and Supplementary Video 1 for the *AII_A-I* group). The issue arises from an upward shift of the elastomer during contact with indenters that have large contact areas (e.g., a hemisphere), caused by inadequate adhesion to the plate. This results in a slight change in the image distribution for a small subset of frames. However, this artifact does not affect the overall force prediction performance for GelSight (A-II) (see **Fig. 4F**).
2. Another failure case arises in heterogeneous translation when the reference marker on a sensor array is too small while the system is overly sensitive to marker size and displacement changes. We recommend choosing the reference marker size and sensitivity parameters to match the marker size and motion characteristics of the source sensor used for transfer. Suggested configurations are provided in Supplementary Text 8 and Supplementary Fig. 12A.
3. In real-time deployment, the force prediction model exhibits zero-shift over the first few seconds and mild shift under static loading. The initial zero-shift can be mitigated by subtracting a baseline, whereas the static shift stems from the elastomer's intrinsic hysteresis issues. Additionally, when grasping metallic objects, the uSkin sensor can fail due to magnetic field interference. Nonetheless, the manipulation tasks are still completed successfully because of the gripper's heterogeneous sensor configuration (see Supplementary Videos 4–7).

Some limitations of our models and possible future work:

1. We still require at least one sensor that is fully calibrated with force-labeled data and material priors to compensate material property. Such a calibrated sensor can be sourced from our dataset or from other existing sensors, followed by collecting a small number of locations-pair reference images. For material prior, instead of using traditional material characterization method, some recent in-situ method [43,44] can be referred to avoid use F/T sensor and moving stage, robot arm. Looking ahead, an important direction is to build foundational tactile database like ImageNet[45] that totally removes the need for any data collection and material calibrations. Given our model's adaptability and data reusability, the dataset is inherently scalable using our data collection trajectory and could enable users worldwide to rapidly equip their own sensors with high-performance force sensing.

2. As our model is based unsupervised learning, it is still not comparable to the accuracy with supervised learning. However, if the user wants to get very high-accuracy force prediction, this model is expected to be very good pretrained model to be finetuned with only a few amounts of force labels by using an existing calibrated sensor.
3. It is still an interesting problem in how to expand this model in transferring between finger-size tactile sensors to large-area electronic skins with extremely thin film and large curved surface, as well as some sensors without explicit taxels such as EIT[46] sensors. We expect this can be solved by dividing the large-area skins into subarea and them transfer in comparable region with finger-sized sensors.

Beyond force sensing, our three key contributions, a unified marker representation, marker-to-marker translation, and a spatiotemporal regression model, can readily extend to broader robotics applications by simply replacing the output layer and the value is vast. These include in-hand pose estimation[47], 3D reconstruction[48], and other tactile tasks critical for dexterous manipulation. By establishing a unified framework for transferable sensing across tactile sensors, our work paves the way for fast, cost-effective, and scalable tactile sensing systems.

# METHODS

## Model derivation

**Problem setting.** In a general case, we have one or more calibrated tactile sensors, referred to as source domains $\{S_i\}_{i=1}^{n}$, $n \in \mathbb{N}$, with paired tactile signal-force data $\{I^{S_i}, F^{S_i}\}$, while uncalibrated tactile sensors, referred to as target domains $\{T_j\}_{j=1}^{m}$, $m \in \mathbb{N}$, are without access to force labels $\{F^{T_j}\}$ for tactile signals $\{I^{T_j}\}$. The problem is how to leverage the collected force labels $\{F^{S_i}\}$ from $\{S_i\}_{i=1}^{n}$ to endow sensors $\{T_j\}_{j=1}^{m}$ with force prediction capability. Notably, the tactile signal can be tactile images or multichannel electrical signals.

**Overview**. Above problem setting raises a cross-sensor force calibration challenge in a many-to-many paradigm regardless of discrepancy in sensing principles, structure designs and material properties. A solution building upon such a problem setting is able to establish a foundation for transferable force sensing across various homogeneous (**Fig. 2**A) or heterogeneous tactile sensors (see **Fig. 2**B). Our proposed GenForce model can be divided into three steps: (1) unified marker representation; (2) marker-to-marker translation; (3) spatiotemporal force prediction.

**Derivation of unified marker representation**. When forces are applied to tactile sensors (**Fig. 1**D), the soft skin deforms and the skin deformation is measured as different types of tactile signals, due to difference in sensing principles. The differences in physical configuration, such as illuminations and cameras in vision-based tactile sensors, or electrode layout and resolution in sensor arrays, lead to further differences in measured signals. To eliminate these input disparities similar to how human mechanoreceptors translate stimuli into sensory encoding, we unify the tactile signals from various tactile sensors as marker representation. Compared with other modality (see Supplemtary Text 1), he marker representation unifies the measured skin deformation as marker deformation and displacement, ensuring a consistent image representation essential for cross-sensor translation by using image translation methods[49,50]. The marker pattern is common in tactile sensors.

In vision-based tactile sensors, it can be deliberately designed on the soft skin, such as GelSight[13], DIGIT[51] or it uses deep learning technique to transfer from markerless skin[52] (see Supplementary Figure 36) While in sensor arrays, the electrodes can also be regarded as another type of marker in either lab prototype[8] or commercial product[15], which can also be measured as marker images from multichannel electrical signals using our signal-to-marker technique (see Fig. 6A, Supplementary Figure 10A-B).

**Derivation of marker-to-marker translation.** By unifying tactile representation as marker-based images, we can transfer the collected data across tactile sensors via image-to-image translation technique at the image level[49,50], as the differences now mainly lie in marker pattern. We enable this transfer across marker images through a marker-to-marker (M2M) translation technique (**Fig. 2**C) that bridges disparities regarding marker size, density, and distribution. To align the image distributions of the source domains $p(I^{S_i})$ with target domains $p(I^{T_j})$, we train an image conditioned diffusion model $G(I_t^{S_i}, I_0^{T_j})$ to map $I_t^{S_i}$ to $I_t^{T_j}$ at deformed time step $t$, $t \in \mathbb{N}$ such that $G: I_t^{S_i} \rightarrow I_t^{T_j}$. The conditional inputs used here are the reference images $I_0^{T_j}$, captured when target sensors are in a non-contact state. Once the model $G$ is trained, $I_t^{S_i}$ can be translated into $I_t^{G_i}$, which matches the image styles of $I_t^{T_j}$ while preserving the deformation information of $I_t^{S_i}$. Note that, we only need to train one single model $G$ available to selectively transfer from sensor $i$ to sensor $j$ by using image conditions.

**Derivation of spatiotemporal force prediction.** Upon marker-to-marker translation, we can construct new datasets $\{(I^{G_i}, F^{S_i})\}$ from existing datasets $\{(I^{S_i}, F^{S_i})\}$ to train force prediction models $\hat{h}_{i \rightarrow j}$ for new sensors $T_j$, where we focus on three-axis force prediction $F = (F_x, F_y, F_z)$ in this paper. Specifically, we leverage force prediction models depicted in **Fig. 2**D to take spatiotemporal information into consideration. In addition, we consider the material hardness effect and formulate a material compensation method (see **Fig. 5**E). As such, force prediction models $\hat{h}_{i \rightarrow j}$ can now be trained with the transferred dataset $\{(I^{G_i}, F^{S_i})\}$ and then predict forces $F^{T_j}$ with images $I^{T_j}$ from new sensors.

## Marker-to-marker translation model

The M2M model (see **Fig. 2**C) comprises two primary components: a marker encoder-decoder [53] and an image-conditioned diffusion model[54]. We adapt the variational autoencoder (VAE) architecture from SD-Turbo[54] as the marker encoder-decoder. The conditional diffusion model is based on the UNet[55] architecture from SD-Turbo combined with a DDPM Scheduler [56]. To optimize the model's performance while maintaining parameter efficiency, we implement Low-Rank Adaptation (LoRA)[57] to key network components, including attention layers, convolutional layers, and projection layers for efficient fine-tuning. The model is mainly trained with one NVIDIA A100 GPU with 80GB memory. We employ an 80-20 train-test split on our dataset. All images are preprocessed from 640×480 pixels to a uniform size of 256×256 and normalized to [0,1] range. We use AdamW optimizer with an initial learning rate of $5 \times 10^{-6}$, betas=(0.9, 0.999), epsilon=$1 \times 10^{-8}$, and weight decay of $1 \times 10^{-2}$. Details of the model architecture and training process are shown in the Supplementary Text 2. The ablation study is available in Supplementary Text 4.

## Spatiotemporal force prediction model

The spatiotemporal architecture has four components (see **Fig. 2**D): (1) a marker feature encoder from RAFT[58] that extracts spatial representations from sequential tactile images, (2) a spatiotemporal module using ConvGRU[58] that captures temporal dependencies while preserving spatial information, (3) a ResNet[59] module that hierarchically processes features through residual blocks, and (4) a regression head with a multilayer perceptron that transforms the processed features into force predictions. We use the same image preprocessing as M2M, while the forces are processed with min-max normalization. We train the model with sequential images with variable length. The training process follows a two-stage approach: first, we pre-train the model on with transferred data with a learning rate of 0.1 for 40 epochs, then fine-tune it with a learning rate of $1\times10^{-3}$ for another 20 epochs. We use SGD optimization with momentum of 0.9 and weight decay of $5\times10^{-4}$. The detailed model architecture and training process are demonstrated in the Supplementary Text 3. The ablation study is available in Supplementary Text 4.

## Simulation of marker deformation

This pipeline (**Fig. 3**A) enables the synthesis of realistic marker deformation images similar to those from real-world sensors. The pipeline comprises three primary stages: (1) skin deformation simulation; (2) marker pattern design; (3) marker deformation rendering. We model the skin deformation simulation environment in Tacchi[60] where sensor dimension, material properties, and contact trajectory) are configured to generate deformed meshes (see Supplementary Figure 1C) under varying locations. The simulation model comprises a $20\times20$ mm elastomer with a thickness of 4 mm. The elastic modulus is set as $1.45\times10^{5}$ Pa with a Poisson's ratio of 0.45. We employ 18 different primitive indenters (Supplementary Figure 1A) following the contact trajectory shown in Supplementary Figure 1B and the Supplementary Text 5 to get deformed meshes. Upon obtaining the deformed meshes, the meshes are imported into Blender for marker image rendering. Before rendering, we draw 12 types of reference marker images (see **Fig. 3**A) with different marker patterns, density, and size using CAD software. For each mesh, we apply the reference marker image as the texture of the deformed mesh and use UV mapping specifically to surfaces with negative z-axis normal (see **Fig. 3**A and Supplementary Figure 1C). The rendering environment is configured with a 35 mm focal length camera, oriented to capture the deformed surface. To ensure visibility of the marker patterns, we set the light strength to 60.0 units. The rendered outputs are $640\times480$ images. In total, we get 810 marker images per marker pattern and a total of 9,720 images across all reference marker images. It is worth noting that this dataset is scalable by adding varieties of marker patterns, sensor dimensions, material properties, and contact trajectories.

## Fabrication of tactile sensors

We primarily utilize two types of vision-based tactile sensors: GelSight[13] and TacTip[12], both fabricated in-house. Related fabrication details for TacTip (palm shape) can be found in [16]. Additionally, we incorporated an off-the-shelf magnetic-based tactile sensor, the uSkin (*Patch* version) from XELA robotics with of a $4\times4$ grid of taxels, where each taxel measures 3-axis deformation. The GelSight sensor integrates a Logitech C270 HD webcam, 3D-printed housing components (fabricated using Bambu Lab P1S), laser-cut acrylic boards ($25\times25$ mm), LED lights with RGB color, and silicone elastomers. To achieve varying illumination conditions, we attach resistors to the RGB lights, producing three GelSight variants with different light colors: sensors with marker Array-II, Circle-II, and Diamond-I patterns. The sensors Array-I and Circle-I use the same hardware as Array-II and Circle-II respectively differing only in marker patterns with slightly random shift and rotation to replicate changes that occur on long-term use. For material compensation testing, we utilized white LEDs. The marker patterns used can be found in

Supplementary Figure 4A. Details for the soft skin fabrication can be found in Supplementary Figure 2A and Supplementary Text 6.

## Data collection in real world

The data collection setup consists of five main components shown in **Fig. 4**A: a UR5e robot arm, a ATI Nano17 force/torque sensor, 3D-printed indenters, tactile sensors, and a 3D-printing support base. Each indenter is mounted on the robot arm with fixed initial positions and orientations. The contact trajectory for the real-world data collection is depicted in Supplementary Figure 2B with parameters in the Supplementary Text 7. We choose five surface points on the skin. For each point, the indenters move towards different target points with varying depths and orientations in four actions: moving down, moving laterally, returning laterally, moving up. The movement speed was set to 25% for GelSight and uSkin sensors, and 40% for TacTip on the UR5e touch panel. For material compensation, we collected approximately 30,000 force-image pairs per skin for force transfer and 3,400 location-paired images per skin for M2M model finetuning. For heterogenous translation, approximately 60,000 force-image paired data are obtained per sensor, including 11,520 location-paired images.

## Marker Extraction

For GelSight and TacTip, we use marker segmentation as illustrated in Supplementary Figure 4Band Supplementary Figure 10C. The marker segmentation process comprises a coarse-to-fine process. For GelSight, we execute rough extraction by combining the thresholding with Gaussian blur. For TacTip, we crop marker area from the raw RGB image and normalize brightness, followed by identifying markers using thresholding. In the fine extraction stage, we employ an EfficientSAM[61] model for both sensors to precisely segment individual markers by using the position information from the rough marks. For uSkin sensor we use signal-to-marker conversion (see Supplementary Figure 10A-B). The raw sensor data contains 16 three-dimensional vectors representing the taxel deformations. We convert the raw signals to marker images by using the parameters in the Supplementary Text 8.

## Material compensation

According to the relationship of force and indentation depth in Supplementary Text 9, we measure the $F-d_z$ relationship for the homogeneous translation (see **Fig.** 5C) and the heterogeneous translation (see Supplementary Figure 13A). For each sensor, we conduct three indentations using the *prism* or cone indenter (Supplementary Figure 13A) with a rigid surface. The indenter is pressed down to a maximum depth $d_{max}$ of 1.4 mm and then released. We collect paired force-depth data during both loading and unloading phases according to the parameters in the Supplementary Text 9. We can then get the fitted lines of the $F-d_z$ curves (see Supplementary Figure 8C, 13A-iii and 13A-iv), using two-degree polynomials. Then, the force label $F^S$ can be corrected to $F^{SC}$ with either the ratio of $r_L$ or $r_U$ depending on the contact is in loading phase L or unloading phase U according to the process in the Supplementary Text 9.

## Robot grasping and slip control

We use a Franka FR3 robot arm with a Franka hand (max opening 0.08 m), an Intel RealSense camera, daily objects (9 for grasping and 4 for slip control), and tactile sensors. For grasping, normal force predicted from GelSight and uSkin feed into a simple proportional force controller. The gripper closes at 0.005 m/s from an object-dependent inital width (about 0.025 m for grapes up to 0.075 m for oranges). Depending on fragility, the target force is 0.6 N, 0.8 N, or 1.2 N (e.g., lower for chips and grapes, higher for rigid items like a meat box). Force controller runs at 10 Hz with Kp=0.0004. The slip-control experiments use the same setup, switching the tactile sensors to TacTip and uSkin. Here we pair the same proportional controller with a top-position slip check: after lifting, we monitor Fx and Fy at 20 Hz for 10 s, keeping a 0.5 s window. If the short-term shear force variation over three samples in either axis exceeds 0.3 N, we tighten the grasp by 0.001 m and command the change at 0.015 m/s. See more details in Supplementary Video 4–7.

## Data Availability

All data for reproducing this work are available in the main text and supplementary material. The collected dataset is available to public upon paper acceptance with a finalized version.

## Code Availability

The code that supports the findings of this study are available to public upon paper acceptance with a finalized version.

**Acknowledgments:** We thank King's College London (2025), King's Computational Research, Engineering and Technology Environment (CREATE) for providing HPC.

**Author contributions:** Z.C. and S.L. conceived the systematic study. Z.C. proposed the model, implemented the code, fabricated the sensors, collected the data, wrote the manuscript and created the figures. J.D. and N.O advised on model implementation and marker segmentation. X.Z. advised on marker simulation and provided TacTip sensor. Z.W., Y.Z., and Y.W. contributed to data collection. E. S. P. funded the Franka FR3 robot arm and supervised experiments regarding robot manipulation. N.L. funded the TacTip sensor and supervised all experiments regarding TacTip. L.J. funded the uSkin sensor and supervised all experiments regarding uSkin. J.D. and S.L. supervised, funded and managed the research, while providing hardware and computing resources. All authors helped revised and polished the manuscript.

**Funding:** This work is supported in part by EPSRC project “ViTac: Visual-Tactile Synergy for Handling Flexible Materials” (EP/T033517/2).

**Competing interests:** The authors declare no competing interests.

**Additional information:**

• AI tools were used solely for grammar checking in this manuscript.

• Supplementary Information & Video is available for this paper.

• Project page is available at https://zhuochenn.github.io/genforce-project/

• Correspondence and requests for materials should be addressed to Shan Luo & Zhuo Chen

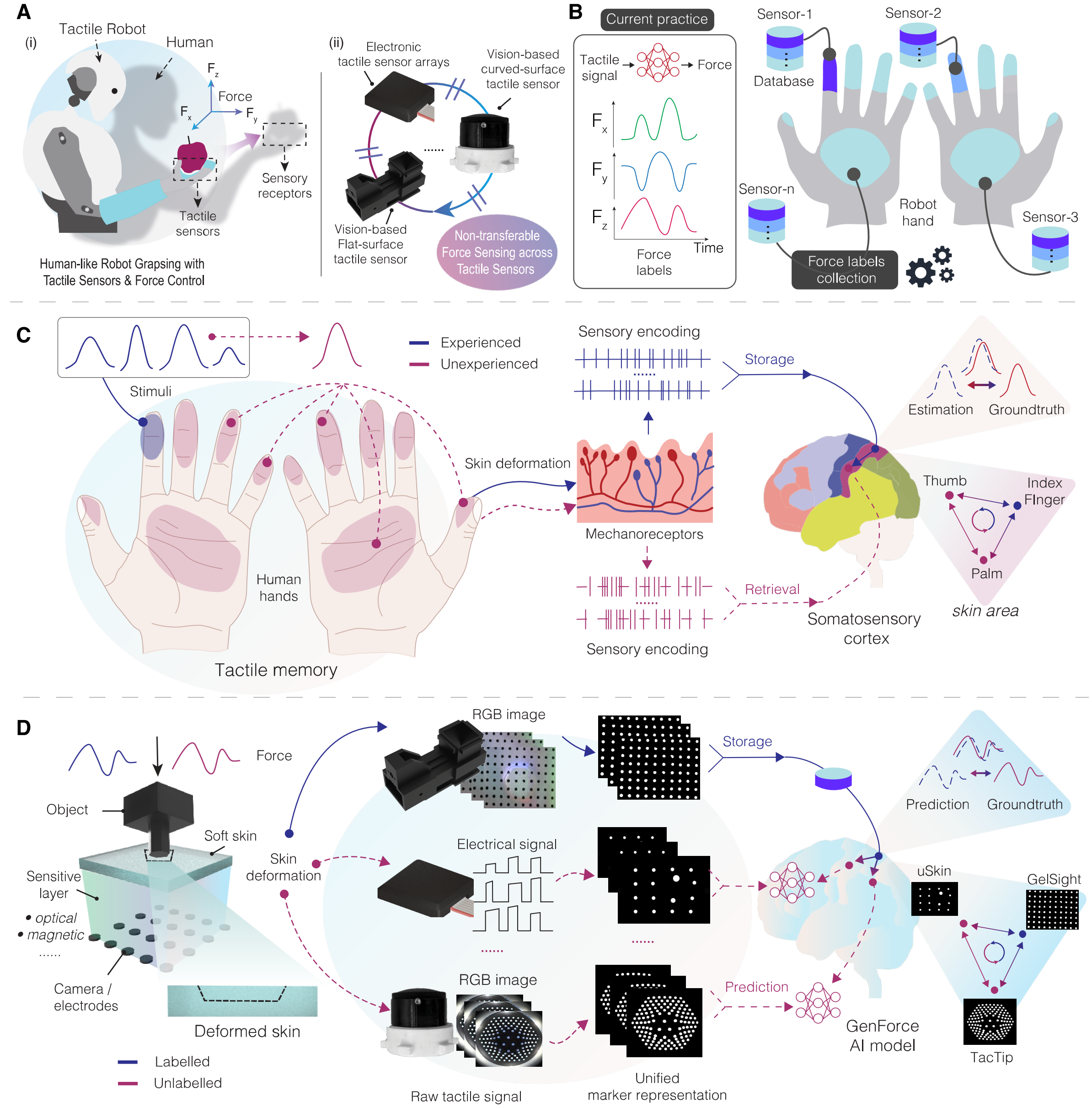

**Fig. 1 | Transferable force sensing. (A)** Robot grasping objects with tactile sensors and force control mimics human actions with sensory receptors. These bio-inspired tactile sensors cannot transfer force data with each other due to differences in sensing principles, structural designs and material properties. **(B)** Current practice to train force prediction models uses repetitive and costly data collection process for force labels. **(C)** Humans use a tactile memory system to estimate stimuli on unexperienced skin area by retrieving tactile memories stored in the somatosensory cortex. **(D)** Overview of the GenForce model. Tactile sensors produce diverse tactile signals under the same deformation due to differences in sensing principles, structural designs and material properties. GenForce unifies tactile signals into marker representation, enables marker-to-marker translation across various sensors, and achieves high-accuracy force prediction on uncalibrated sensors using data transferred from calibrated sensors.

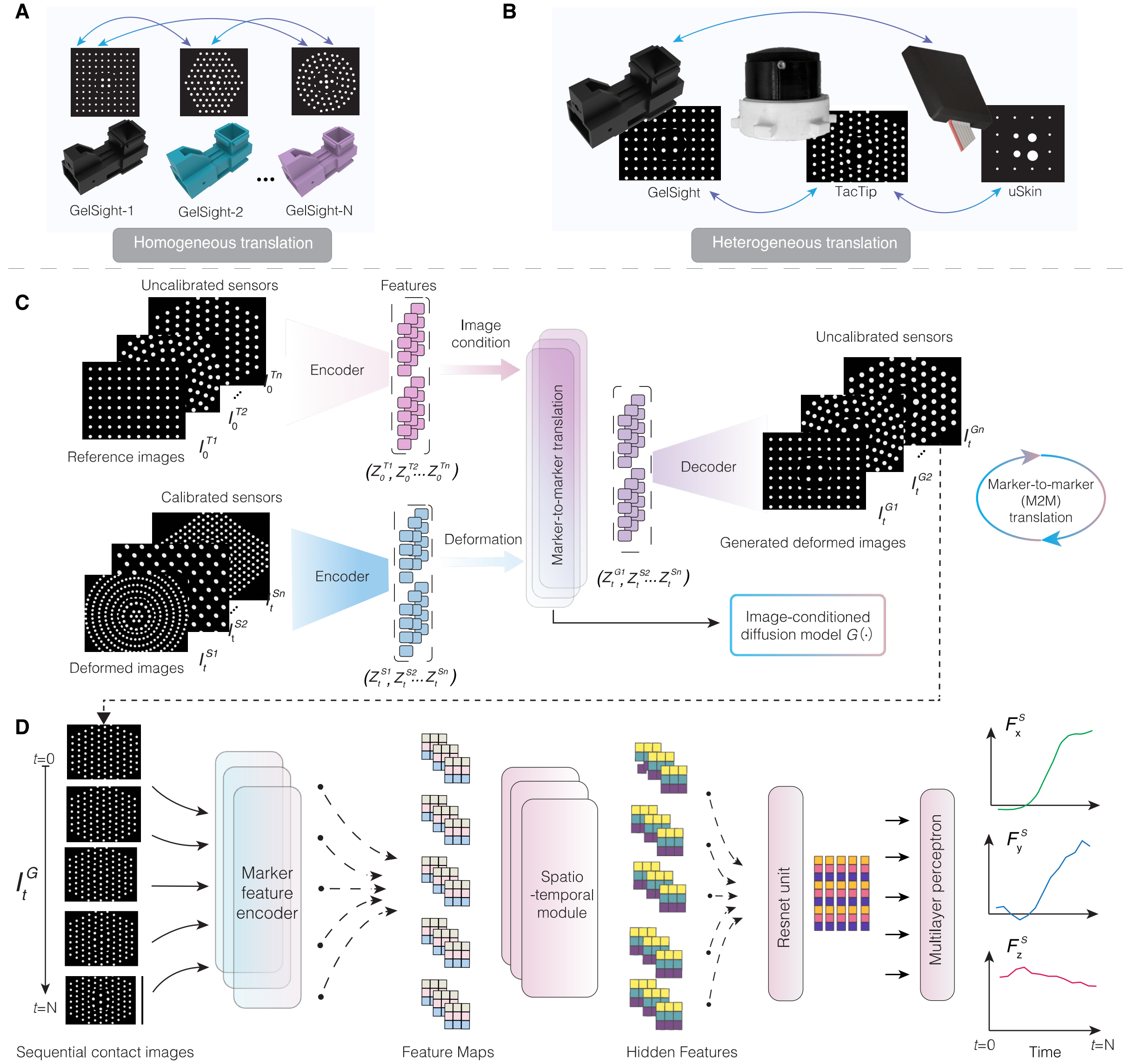

**Fig. 2 | Marker-to-marker translation and spatiotemporal force prediction. (A)** Homogeneous translation demonstrated with GelSight [13] sensors featuring varying marker patterns. **(B)** Heterogeneous translation demonstrated with GelSight, TacTip [16], and uSkin [15] sensors. **(C)** Marker-to-marker translation (M2M) model. The M2M model uses deformed images from calibrated sensors as input and reference images from uncalibrated sensors as conditions to generate deformed images that mimic the deformation applied to uncalibrated sensors. **(D)** Spatiotemporal force prediction model. This model takes sequential contact images as input and outputs three-axis forces, enhancing prediction accuracy through a spatiotemporal module.

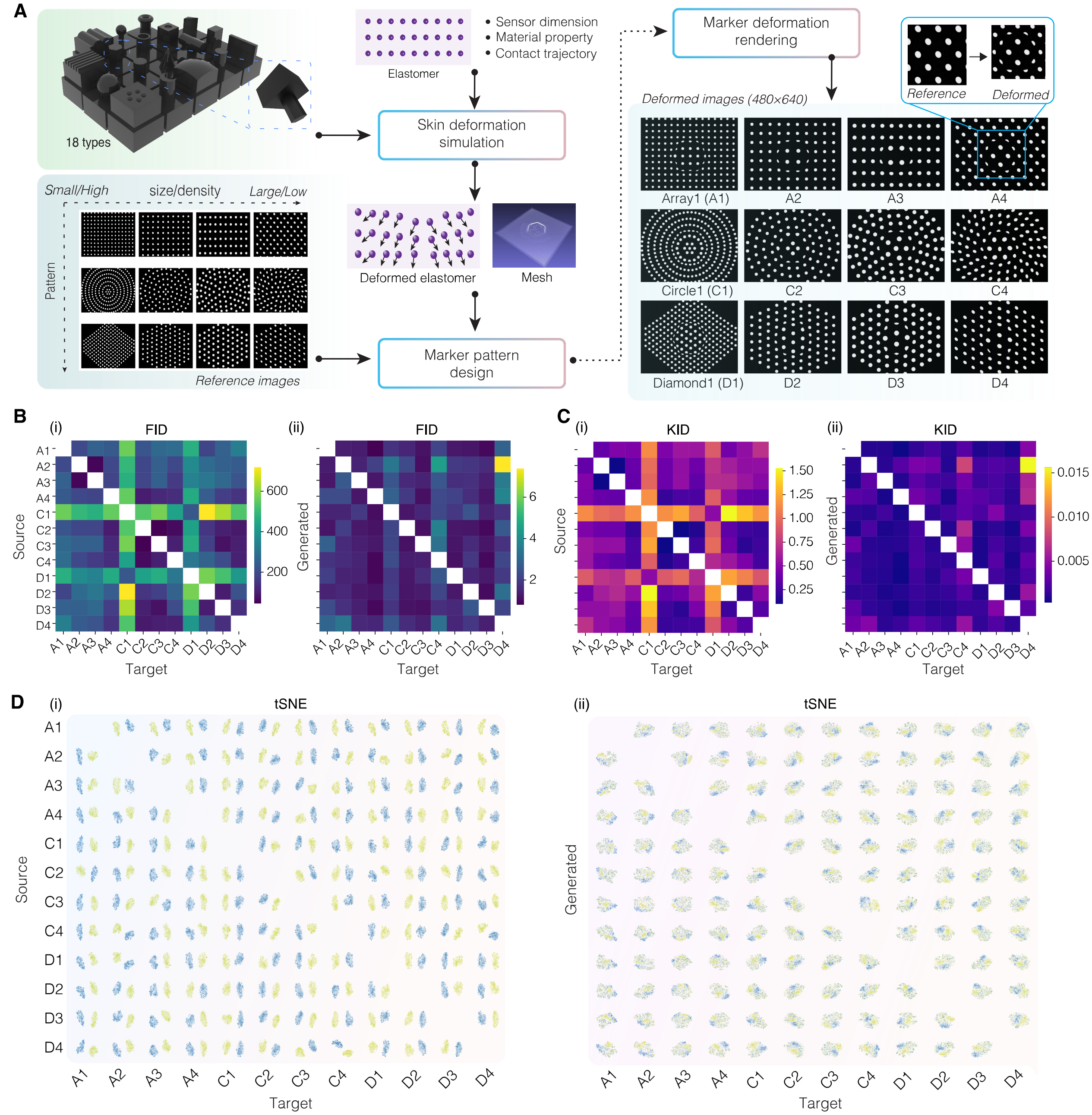

**Fig. 3 | Marker-to-marker translation in simulation. (A)** The pipeline for marker deformation simulation. **(B-C)** Quantitative evaluation of the marker-to-marker translation is presented using heatmaps of FID (B) and KID (C), comparing performance before (i) and after (ii) applying the M2M model on all 132 sensor combinations. **(D)** Feature space visualization using t-SNE illustrates the alignment of target images and generated images across all sensor combinations before (i) and after (ii) applying the M2M model.

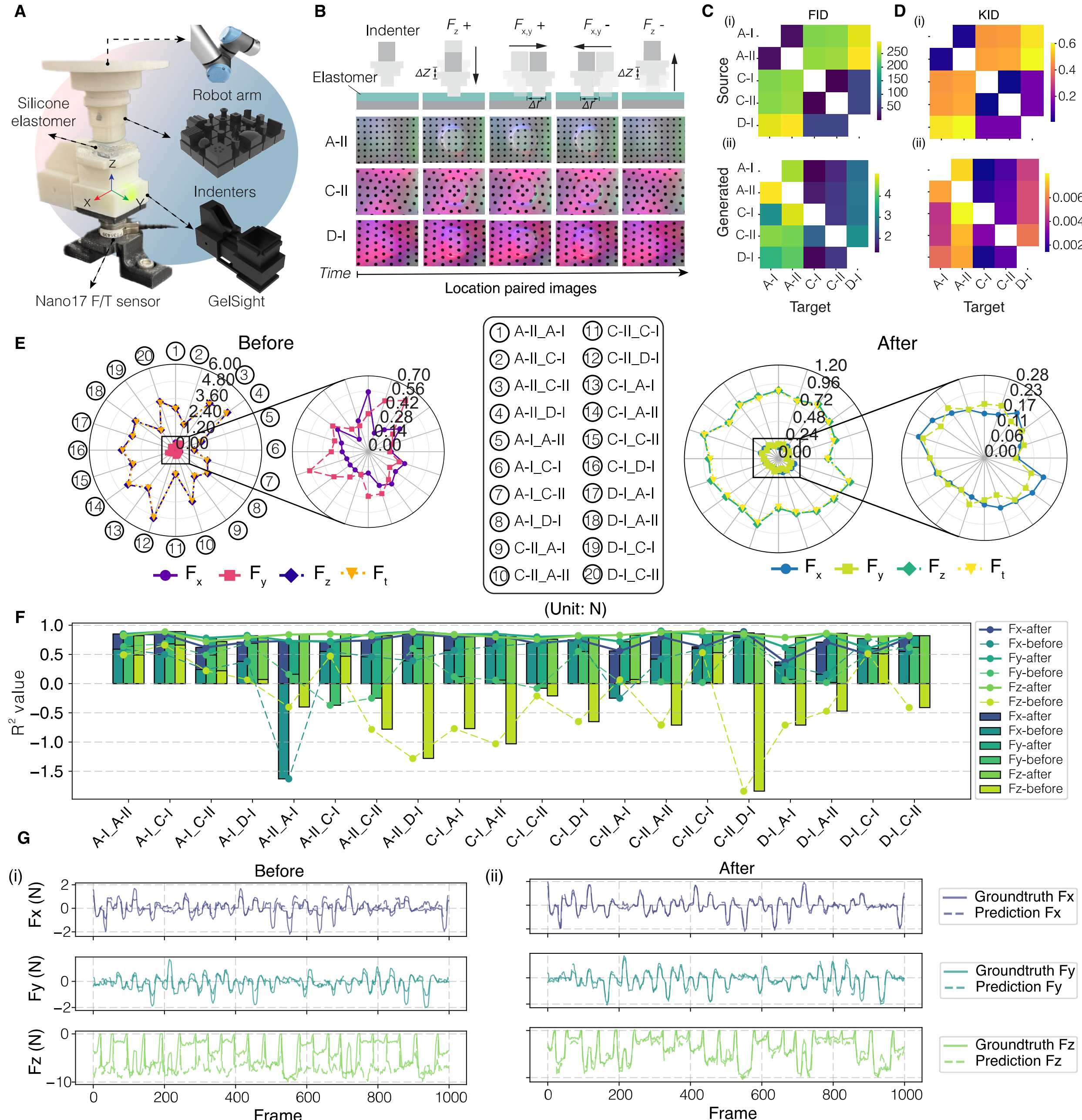

**Fig. 4 | Homogeneous tactile force translation. (A)** Data collection setup. (**B)** Data collection trajectory for acquiring sequential tactile images under normal and shear forces. **(C-D)** Heatmaps of the FID (**C**) and KID (**D**) results before (i) and after (ii) using the M2M model. **(E)** Comparison of force prediction errors before (i) and after (ii) using GenForce model in radar plots. $F_t$ denotes the total force. All units are Newton (N). **(F)** Histogram of the coefficient of determination ($R^2$) values before and after using GenForce model. **(G)** Real-time force prediction comparison over 1000 frames before (i) and after using GenForce model (ii) in group *A-II_D-I*.

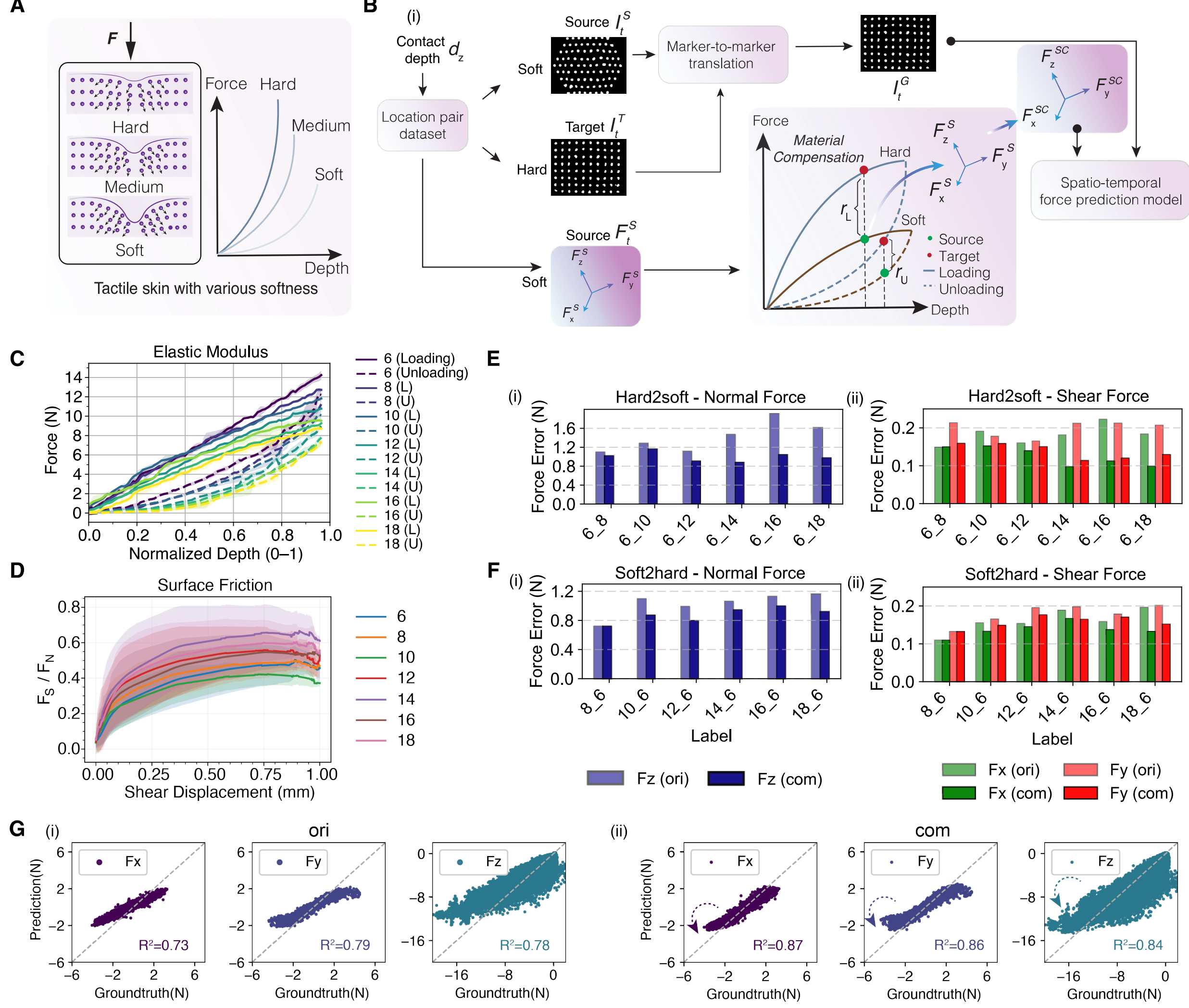

**Fig. 5 | Material hardness effect and compensation. (A)** Schematic representation of tactile skins with varying hardness and their corresponding force-depth relationships. **(B)** Material compensation process incorporates material priors to correct force labels during loading and unloading phases. **(C)** Measured force-normalized depth curves for the seven elastomers during loading and unloading phases, demonstrating elastomers' hardness and hysteresis property. **(D)** Relationship between the shear-to-normal force $F_s / F_N$ and shear displacement of seven elastomers measured across different contact points by using our data collection trajectory. **(E-F)** Force prediction errors (i-ii) before and after using material compensation in *hard-to-soft* group (E) and *soft-to-hard* group (F). **(G)** Fit of force prediction to ground truth demonstrates the effectiveness of material compensation when transferring from a sensor with *r18* (soft) skin to *r6* (hard) skin before (i) and after (ii) using material compensation.

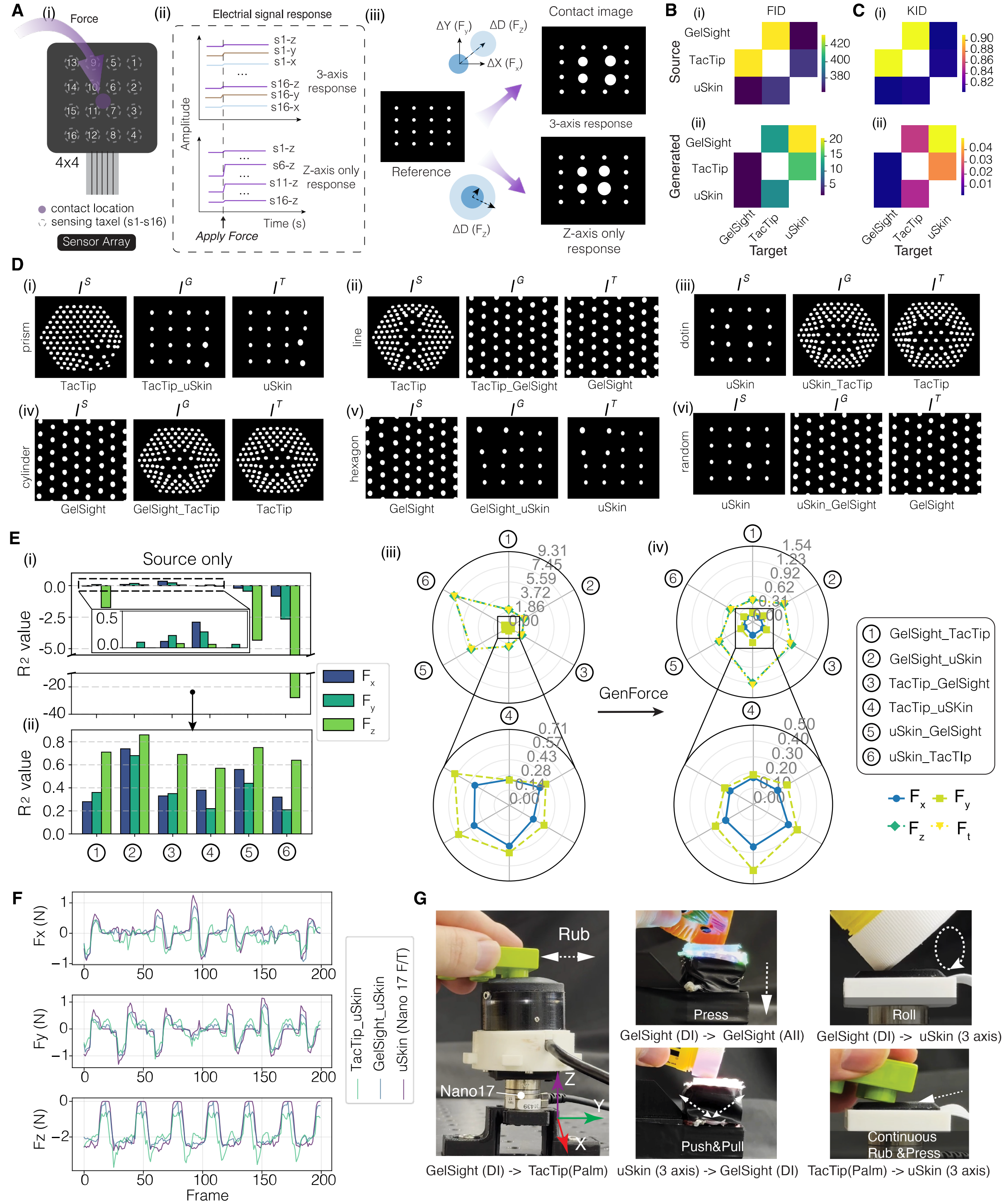

**Fig. 6 | Heterogeneous tactile force translation. (A)** Signal-to-marker processing. The multichannel electrical signals (ii) from a sensor array (i) can be unified to marker images (iii) regardless of 3-axis sensing capability or z-axis only sensing capability. **(B-C)** FID (D) and KID **(E)** before (i) and after (ii) using GenForce model. **(D)** Examples of tactile images after M2M translation. **(E)** Histogram of $R^2$ values (i-ii) and radar plot (iii-iv) of MAE before using GenForce and after using GenForce with material compensation. **(F)** Real-time demonstration of force prediction when target domains is uSkin (3-axis) after using GenForce model with material compensation. **(G)** Force prediction with dynamic contact events compared with ATI nano 17 F/T sensor. Also see animated version in Supplementary Video 2 and 3.

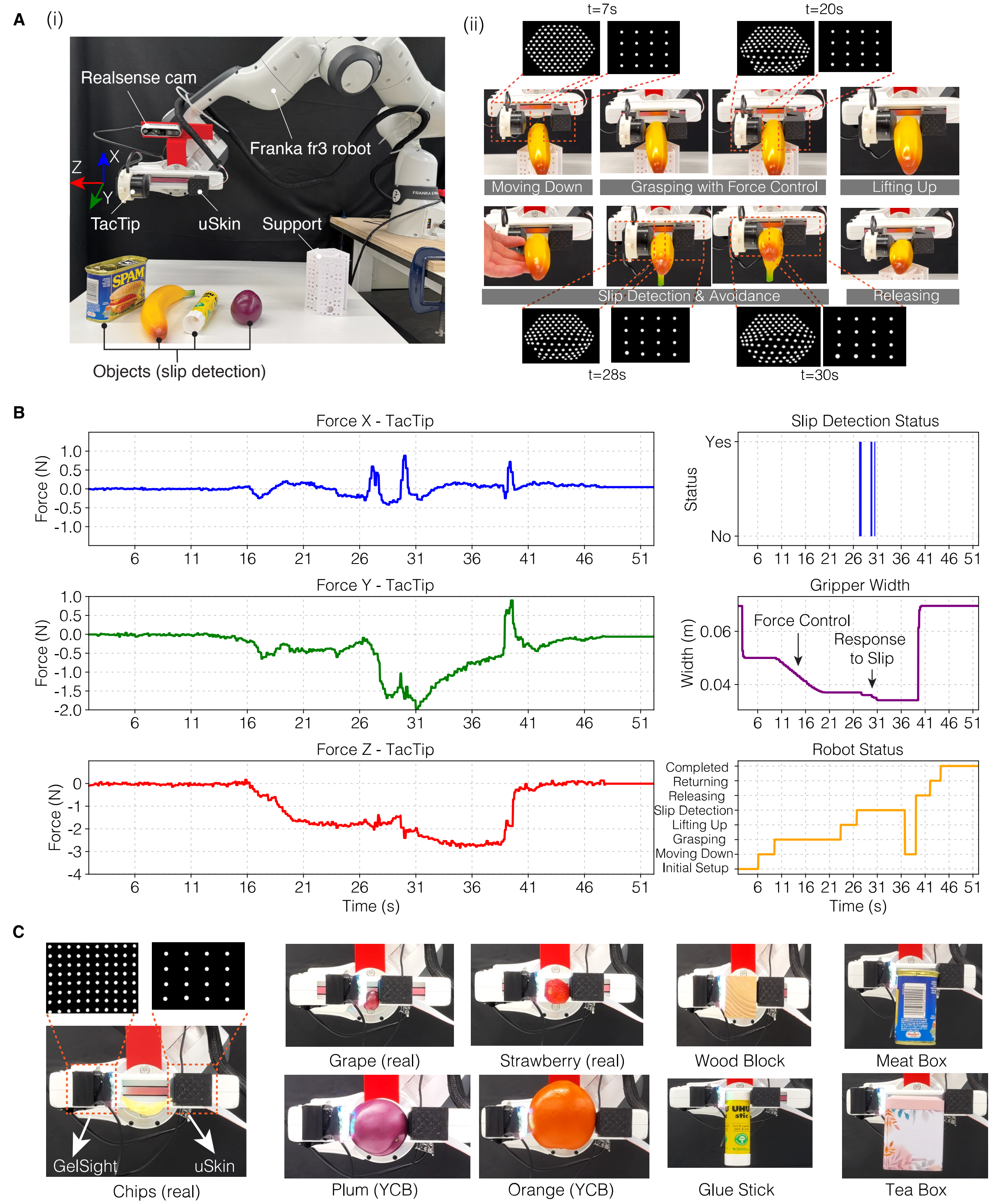

**Fig. 7 | Robot grasping and slip avoidance with transferable force sensing. (A)** Transferable force sensing in robot slip detection and avoidance. (i) Robot arm equipped with TacTip (palm) and uSkin (3 axis). (ii) Sequence of robot grasping an object with force control, slip detection, and avoidance. **(B)** Real-time measurement of forces, slip-detection status, gripper width, and robot status during grasping task of A-ii. **(C)** Daily objects grasping with transferable force sensing and control using GelSight (*A-II*) and uSkin (3 axis). All forces models used in each sensor are transferred from other sensors. See more details in our Supplementary Video 4-7.